\documentclass[conference]{IEEEtran}
\IEEEoverridecommandlockouts
\usepackage{cite}
\usepackage{amsmath,amssymb,amsfonts}
\usepackage{algorithmic}
\usepackage{graphicx}
\usepackage{textcomp}
\usepackage{xcolor}
\usepackage[utf8]{inputenc} 
\usepackage[T1]{fontenc}    
\usepackage{hyperref}       
\usepackage{url}            
\usepackage{booktabs}       
\usepackage{nicefrac}       
\usepackage{microtype}      
\usepackage{lipsum}
\usepackage{fancyhdr}       
\usepackage{graphicx}       

\usepackage{comment}
\usepackage{siunitx}
\usepackage{subfig}
\usepackage{bm}
\usepackage{enumitem}
\usepackage{tabularx}
\usepackage{longtable}
\usepackage{orcidlink}
\usepackage{cite}
\usepackage{pifont}
\newcommand{\cmark}{\text{\ding{51}}}
\newcommand{\xmark}{\text{\ding{55}}}

\usepackage{xspace}
\newcommand{\ie}{i.e.\@\xspace}

\newcommand{\figref}[1]{Figure~\ref{#1}}

\begin{document}

\title{
Human-Robot Interface for Teleoperated Robotized Planetary Sample Collection and Assembly
}

\author{\IEEEauthorblockN{Lorenzo Pagliara\textsuperscript{*}, Vincenzo Petrone\textsuperscript{*}, Enrico Ferrentino and Pasquale Chiacchio}
\IEEEauthorblockA{Department of Computer Engineering, Electrical Engineering and Applied Mathematics (DIEM) \\
University of Salerno\\
84084 Fisciano, Italy \\
e-mail: \{lpagliara, vipetrone, eferrentino, pchiacchio\}@unisa.it }
\thanks{\textit{\textsuperscript{*} L. Pagliara and V. Petrone are co-first authors}}
\thanks{Accepted version, published at \url{https://doi.org/10.1109/MetroAeroSpace57412.2023.10189984} --- 2023 IEEE 10th International Workshop on Metrology for AeroSpace (MetroAeroSpace)}
\thanks{\textcopyright2023 IEEE.
Personal use of this material is permitted.
Permission from IEEE must be obtained for all other uses, in any current or future media, including reprinting/republishing this material for advertising or promotional purposes, creating new collective works, for resale or redistribution to servers or lists, or reuse of any copyrighted component of this work in other works.}
}

\maketitle

\begin{abstract}
As human space exploration evolves toward longer voyages farther from our home planet, in-situ resource utilization (ISRU) becomes increasingly important. Haptic teleoperations are one of the technologies by which such activities can be carried out remotely by humans, whose expertise is still necessary for complex activities. In order to perform precision tasks with effectiveness, the operator must experience ease of use and accuracy. The same features are demanded to reduce the complexity of the training procedures and the associated learning time for operators without a specific background in robotic teleoperations. Haptic teleoperation systems, that allow for a natural feeling of forces, need to cope with the trade-off between accurate movements and workspace extension. Clearly, both of them are required for typical ISRU tasks. In this work, we develop a new concept of operations and suitable human-robot interfaces to achieve sample collection and assembly with ease of use and accuracy. In the proposed operational concept, the teleoperation space is extended by executing automated trajectories, offline planned at the control station.
In three different experimental scenarios, we validate the end-to-end system involving the control station and the robotic asset, by assessing the contribution of haptics to mission success, the system robustness to consistent delays, and the ease of training new operators.
\end{abstract}


\section{Introduction}

Many space agencies and organizations around the world have a long-term objective of sending humans into deep space to explore destinations like Mars and beyond. Such a challenging goal requires intensive study in terms of evaluating possible scenarios, strategies, architectures, and mission elements \cite{viscio_methodology_2013}, as well as solid international cooperation and collaboration. An innovative effort in this direction is NASA's Artemis program \cite{nasa_artemis_2022}, which by 2024 plans to land the first woman and the next man on the lunar South Pole, establishing a sustainable infrastructure on the surface \cite{nasa_lunar_2022} and in the lunar orbit \cite{mars_gateway_2019, coderre_concept_2019}. This will enable the crew to spend more time and explore more of the Moon than ever before, gaining invaluable knowledge and experience for future human missions. 

\begin{figure}[!t]
\centering
    \subfloat[Pre-collection \label{fig:pre-grasp-pose}]{\includegraphics[width=0.325\columnwidth]{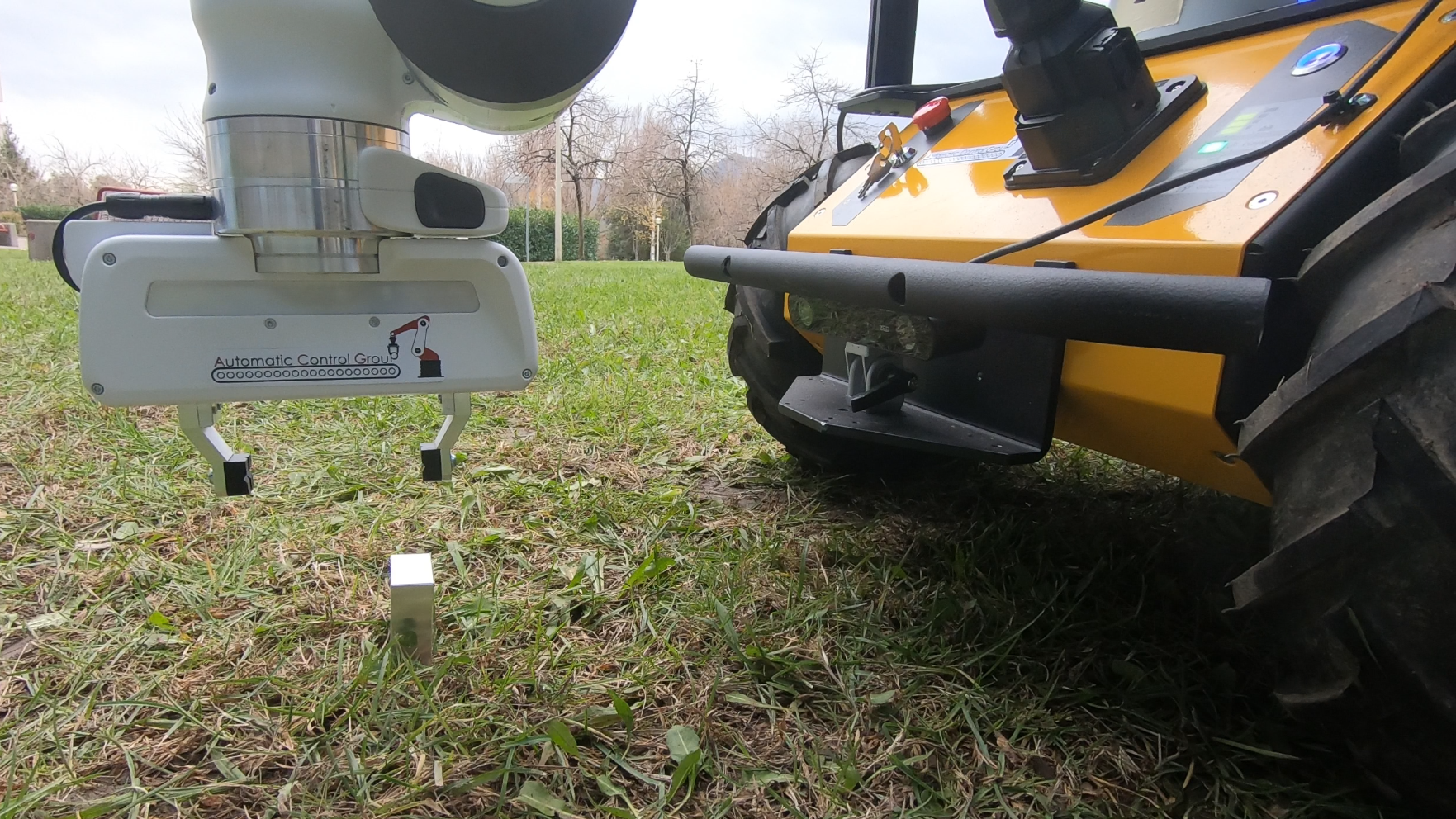}}
    \hfill
    \subfloat[Resource collection \label{fig:grasp-pose}]{\includegraphics[width=0.325\columnwidth]{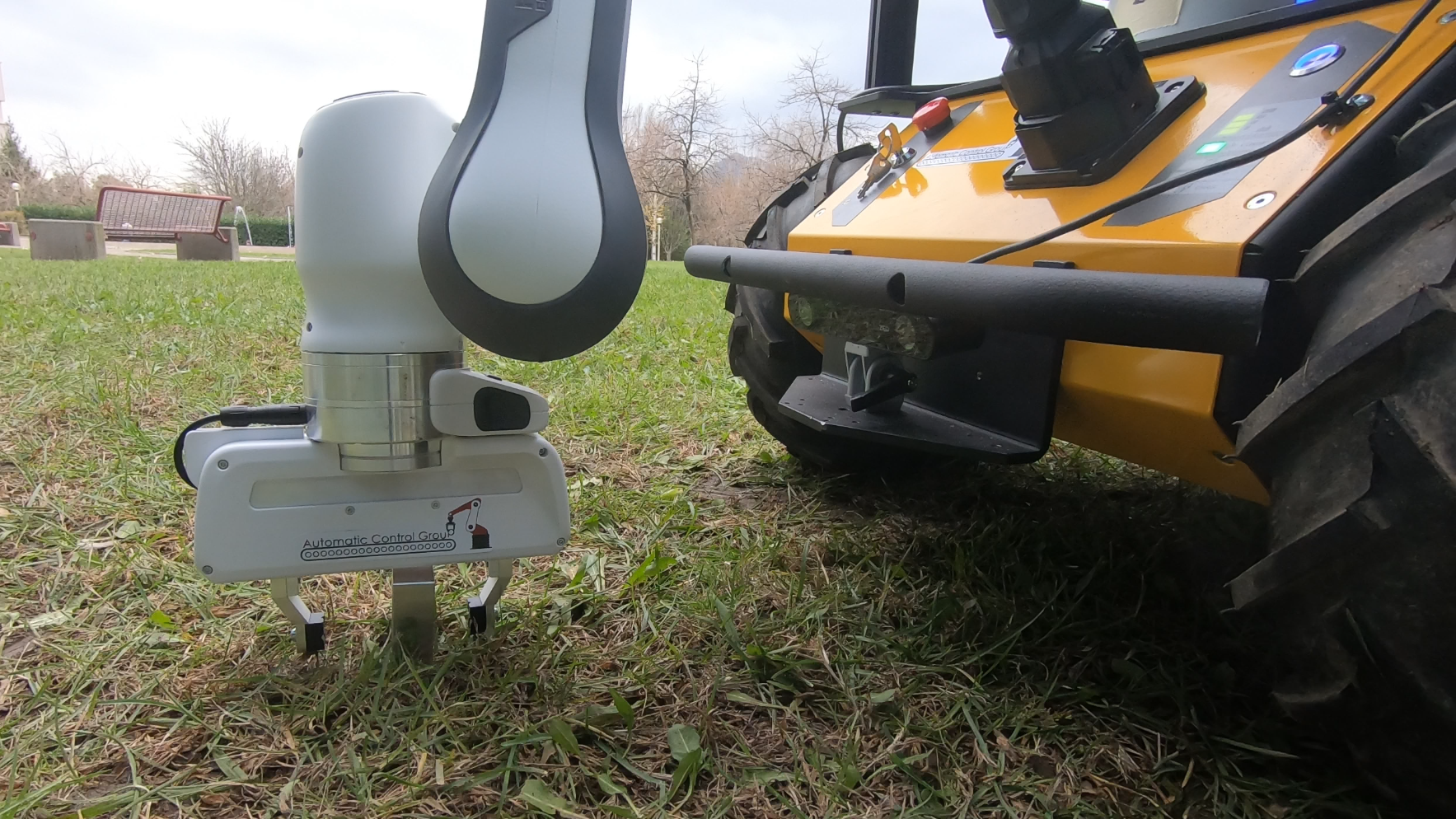}}
    \hfill
    \subfloat[Post-collection \label{fig:grasp}]{\includegraphics[width=0.325\columnwidth]{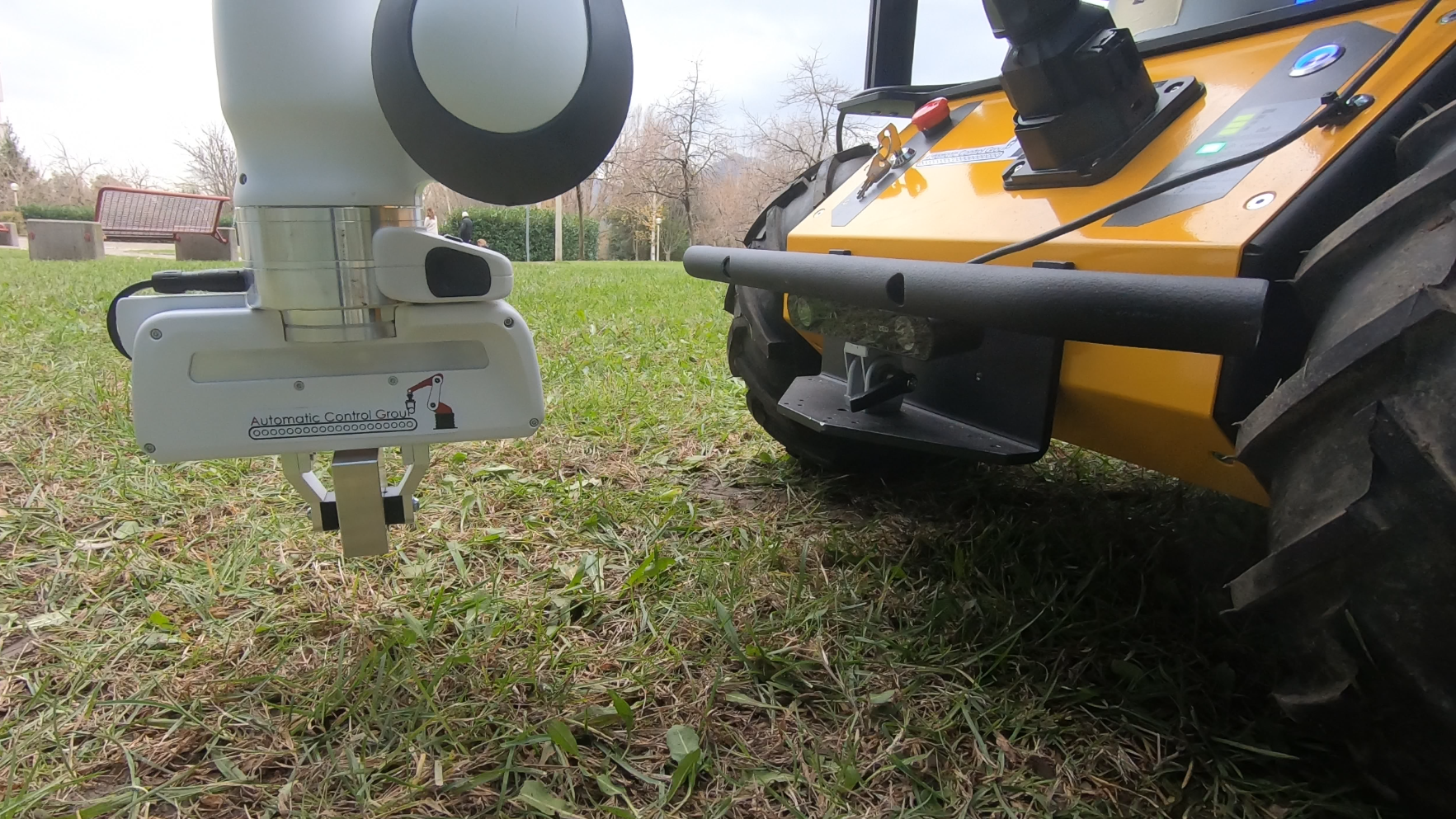}} \\
    \subfloat[Pre-utilization \label{fig:pre-assembly-pose}]{\includegraphics[width=0.325\columnwidth]{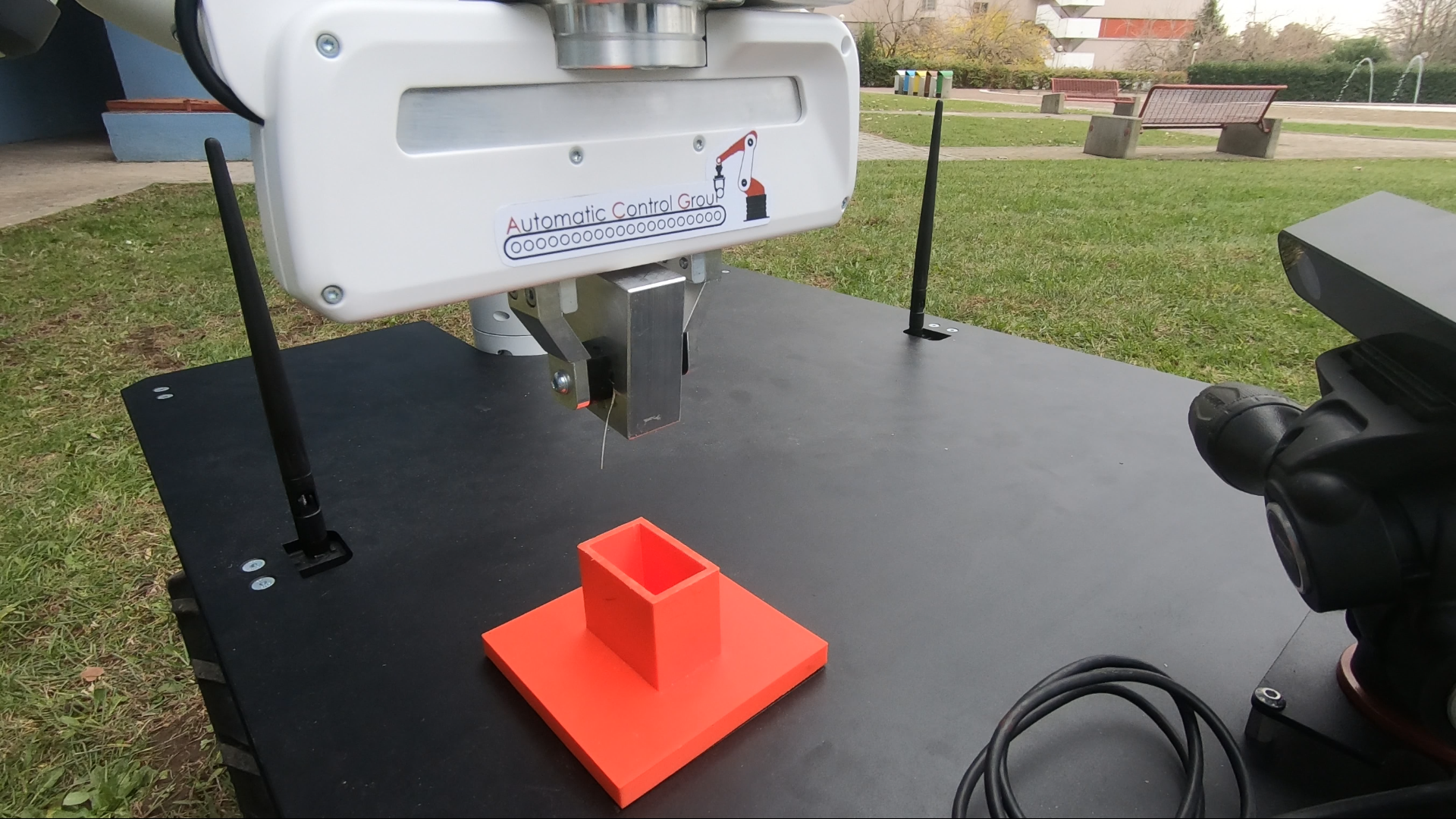}}
    \hfill
    \subfloat[Resource utilization \label{fig:peg-in-hole}]{\includegraphics[width=0.325\columnwidth]{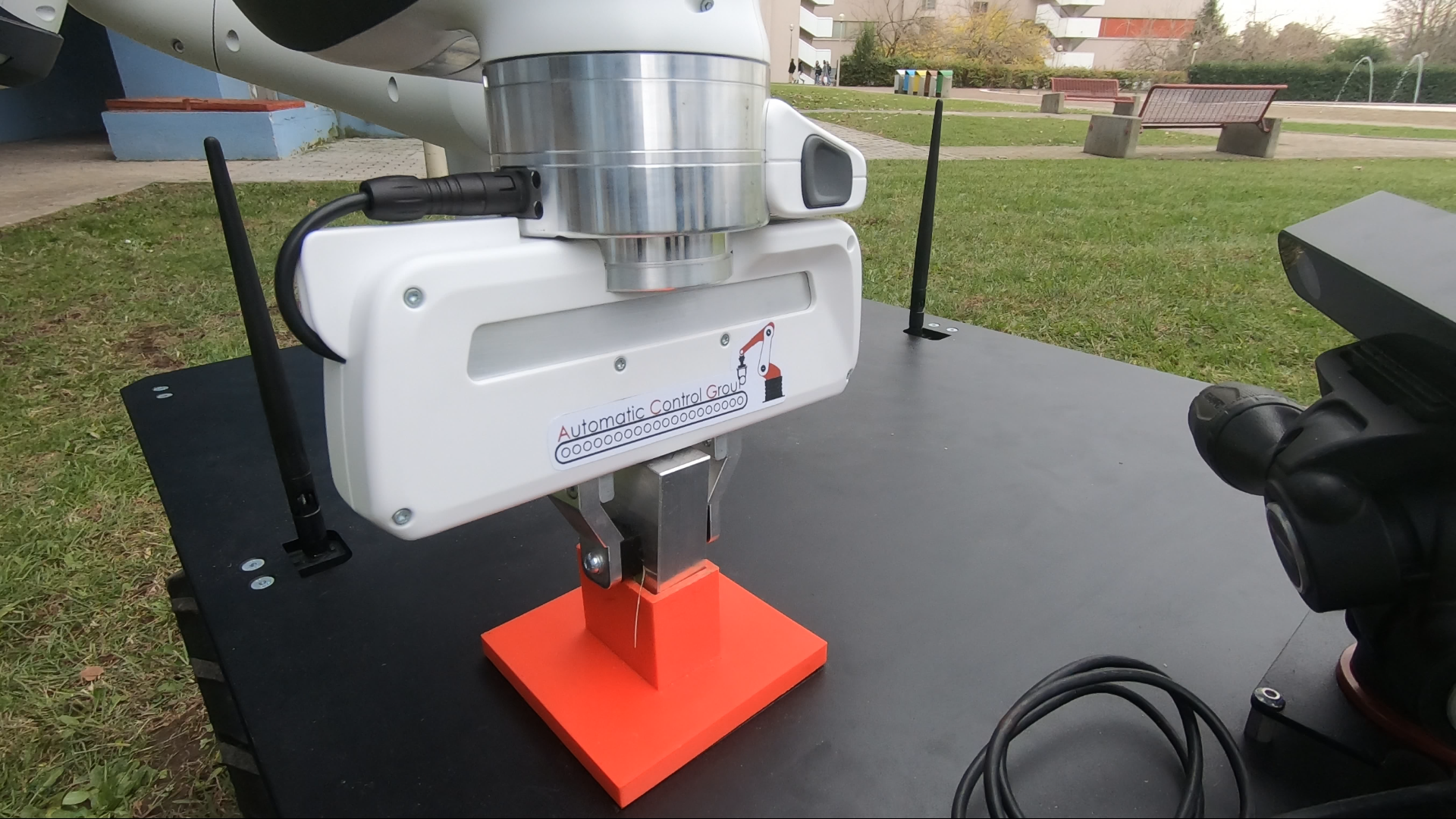}}
    \hfill
    \subfloat[Post-utilization \label{fig:release}]{\includegraphics[width=0.325\columnwidth]{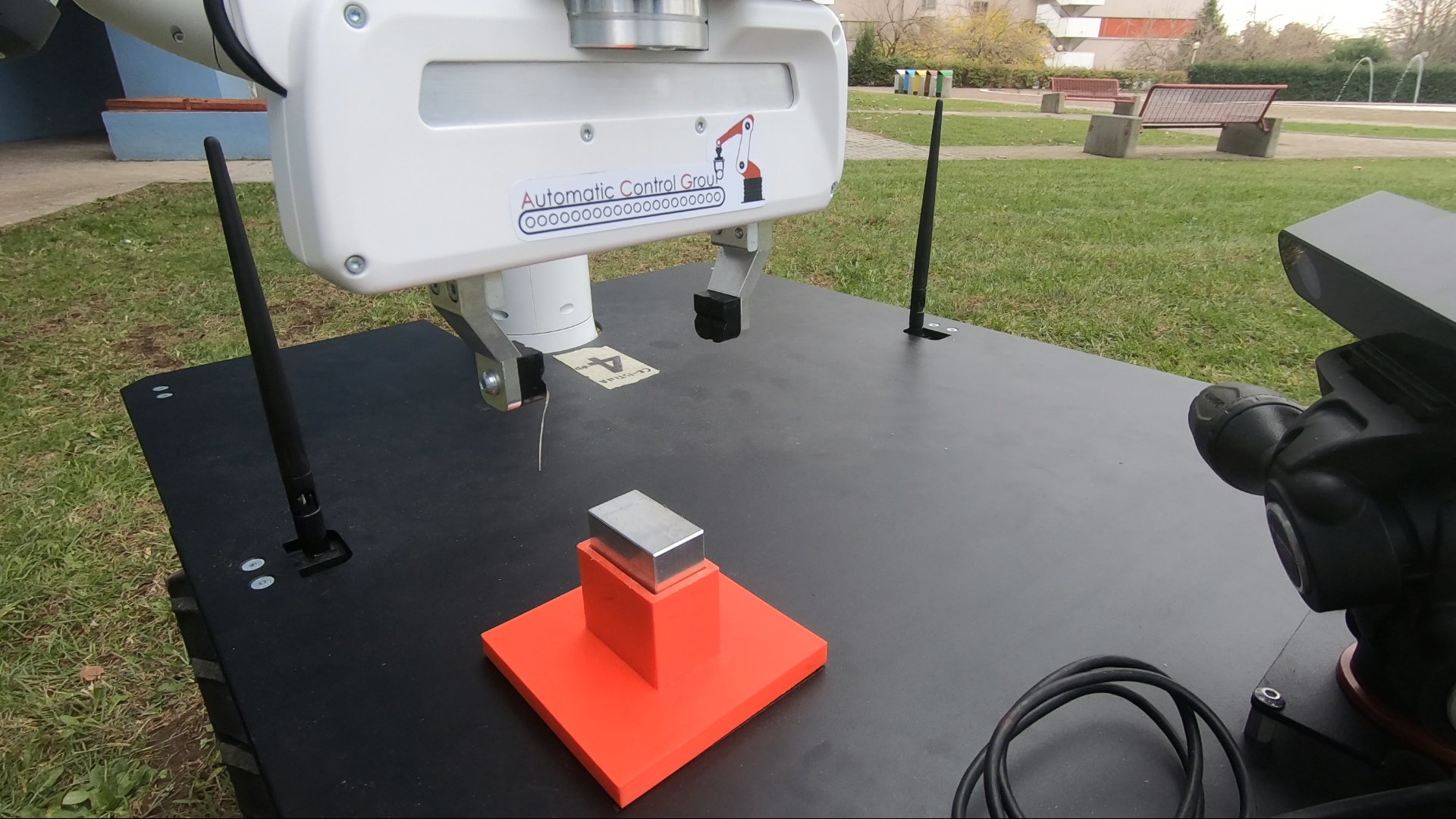}}    
\caption{Example of ISRU mission operations pipeline \cite{ferrentino_autonomous_2022}}
\label{fig:experimental-pipeline}
\end{figure}

One key aspect of space exploration and colonization is the use of resources that are available on other celestial bodies, a practice defined as in-situ resources utilization (ISRU), an example of which is shown in \figref{fig:experimental-pipeline}. The use of in-situ resources can significantly reduce the cost of logistical challenges of space exploration and colonization by allowing for local production of rocket fuel, extraction of water, and mining of minerals and other materials for use in construction and manufacturing.

Accessing, extracting, and using such resources can be particularly dangerous for humans because of the hostile environments on the surface of extraterrestrial planets. For this reason, a common solution for achieving such goals is the use of telerobots. They have been used since the earliest space missions and allowed performing a range of works to increase the productivity of space exploration. Because of communication delays, space agencies have been working to develop robots that can be remotely controlled on the surface of planets by astronauts in orbiting spacecrafts rather than ground stations on Earth.

First, NASA and ESA, with the HET and METERON projects, respectively, conducted teleoperation experiments of robots on Earth by astronauts orbiting in the ISS \cite{bualat_preparing_2014}. The METERON Haptics experiments investigated the effects of microgravity on haptic feedback perception using a 1-DOF force-feedback joystick to teleoperate a robot on Earth from the ISS \cite{schiele_haptics-1_2016}. The experiments demonstrated the absence of significant alterations in flight when compared to ground data, hence the feasibility of bilateral control with force feedback, with time delays in the order of \SI{820}{\milli\second} \cite{schiele_haptics-2_2016}. Using the same experimental setup, the METERON Interact experiment employed force-feedback teleoperation to complete a sub-millimeter peg-in-hole task \cite{schiele_towards_2015}. During the experiment, the astronaut was supported by visual markers and predictive video information of pending commands.

During the experiments of the KONTUR-2 mission \cite{weber_teleoperating_2019, artigas_kontur-2_2016, stelzer_software_2017}, astronauts used a 2-DOF force-feedback joystick to control surface robots and feel the forces of interaction with the environment, using an Earth-ISS communication link characterized by a latency of 20--30~\si{\milli\second}. The bilateral control, based on the Time Domain Passivity Control approach, used during the experiments, enabled stability and performance even in the presence of jitter and data losses, demonstrating the effectiveness of haptic teleoperation with force feedback for deploying robots in prior unknown situations, even with delays up to \SI{1}{\second}.

More recently, during the Analog-1 experiments, a mobile manipulator was commanded on Earth, via a 6-DOF force-feedback haptic device, to complete a rock-collecting task \cite{panzirsch_exploring_2022}. The experiment demonstrated the effectiveness of haptic telemanipulation even with a constant communication delay of about \SI{850}{\milli\second}. 

Parallel to the investigation of haptic teleoperations, METERON SUPVIS-E and METERON SUPVIS-M experiments explored supervised autonomy as a modality of telerobotic control. In this context, the operational concept was based on the employment of intuitive GUIs to perform teleoperation with task-level commands \cite{schmaus_preliminary_2018, schmaus_continued_2019}. This concept offers two main advantages: it ensures the reliability of teleoperations even in the presence of extreme delays in the communication link, and reduces the physical and mental workload of astronauts during teleoperation phases, significantly facilitating the training of new inexperienced operators \cite{schmaus_knowledge_2020}. More recent developments of this operational concept integrated haptic teleoperations as a modality of robot commanding, allowing astronauts to complete complex tasks where human cognitive capabilities and operational flexibility and dexterity are crucial \cite{schmaus_realizing_2022}.

Carrying out such activities effectively requires expertise and long training sessions for astronauts \cite{steimle_astronaut_2013}. In particular, missions with robotic co-workers often require additional training of astronauts, often supported by hardware-in-the-loop VR simulation systems \cite{garcia_training_nodate}. Motivated by this, in the present work, we specifically target those elements of typical teleoperation systems that negatively affect the ease of learning and use. Then, we propose a new operational concept and associated Human-Robot Interface (HRI) to overcome such limitations.

In haptic teleoperations, usability is mainly affected by command mapping between the master haptic device and the slave robot. In position mapping, 1:1 (or lower) scaling is paramount for accurate operations, but the robot's task space is confined to be not larger than the haptic device's, which limits the manipulator's reachability. On the other hand, mapping the haptic device's velocity, or adopting a workspace-extending position scaling would negatively affect accuracy and ease of use, both for precision and large movements. In this work, we tackle this trade-off by designing a set of software tools, supporting an operational concept where the robot task space, subject to 1:1 position mapping, is extended through offline-planned point-to-point trajectories. Therefore, through the alternation of haptic teleoperations and autonomously planned collision-aware trajectories, we aim at facilitating the accomplishment of mission goals, simplifying training procedures, and reducing the learning time for new operators without previous experience with robotic teleoperations. Supported by quasi-real-time visual and force feedback, as well as suitable planning tools integrated into an all-encompassing HRI, the operator can complete our validation objectives with ease, which consist in fetching a resource from the planetary soil and assembling it on board.

\section{HRI for haptic control in ISRU missions} \label{sec:experimental-setup}

\subsection{System Design} \label{sec:system-overview}

\begin{figure}[!t]
\centering
\includegraphics[width=\columnwidth]{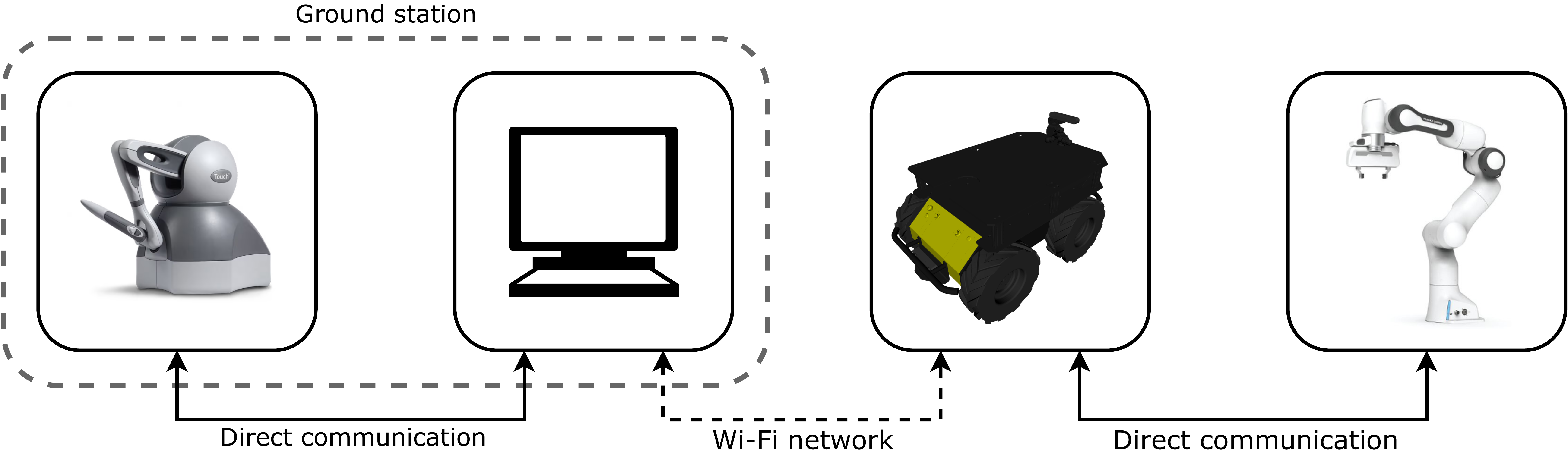}
\caption{High-level overview of the system architecture}
\label{fig:hardware-architecture}
\end{figure}

We assume a typical system setup as in \figref{fig:hardware-architecture}, including a control station and two robots, which we term Exploration Robot (ER) and Manipulation Robot (MR), with the latter mounted on the former. Our MR is a Panda arm by Franka Emika \cite{franka_emika_franka_2022} equipped with a two-finger gripper: it is the entity that physically interacts with the resource, actually performing the sample collection and assembly tasks. Our ER is a customized version of Husky by Clearpath Robotics \cite{clearpath_robotics_husky_2022}. It is a wheeled Unmanned Ground Vehicle (UGV): it serves as an exploration agent that drives toward the resource to manipulate. It is provided with cameras framing the points in which fetching and assembly operations are performed.

The MR is subject to Cartesian impedance control. First, it guarantees stability, even in case of communication loss since, in case of missing references, the controller keeps tracking the last position received. Second, it prevents damage to the arm structure or the manipulated sample, as it adapts the received references to ensure safe interaction with the surroundings \cite{mason_compliance_1981}.

The control station features a workstation and a haptic device \cite{3d_systems_touch_2016}, with a stylus at its tip, installed in a laboratory, with the robots located outdoors. The control station communicates over a Wi-Fi network with the ER, which accounts for forwarding the commands to the MR through its onboard computer.

\begin{figure}[!t]
\centering
\includegraphics[width=\columnwidth]{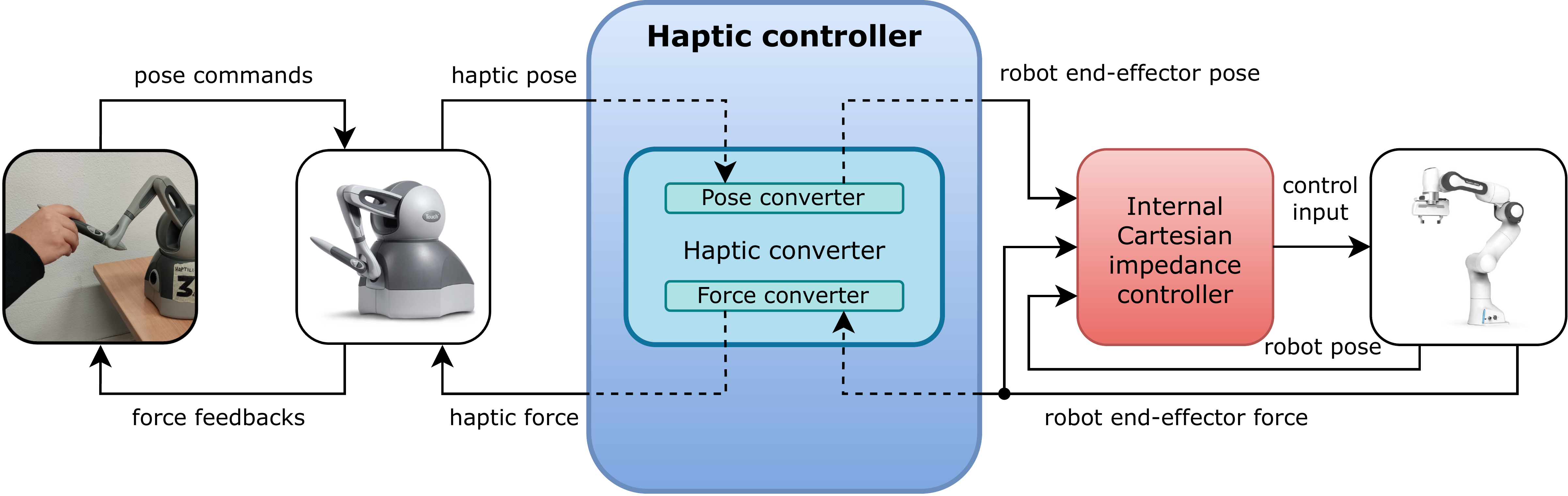}
\caption{Haptic Control System block scheme}
\label{fig:haptic-controller}
\end{figure}

The proposed HRI is made of two components, namely Haptic Control System (HCS) and Robot Visualization \& Planning (RVP), whose name is inspired by \cite{fadrique_exomars_2018}. HCS establishes the mapping between the haptic device stylus' and robot end-effector's poses and renders at the stylus the force feedback perceived at the MR's flange so that they can be felt by the human operator. Its logic is illustrated in \figref{fig:haptic-controller}, where arrows indicate the data flow between the human, the robot, and the haptic device. RVP supports autonomous planning and validation of trajectories and manages transitions from autonomous mode to haptic teleoperations. Their goals and functions are explained in Section \ref{sec:haptic-control-system} and Section \ref{sec:robot-visualization-for-remote-planning-and-motion-control}, respectively, after introducing our mission operations concept in Section \ref{sec:experiment-pipeline-description}.

\subsection{Mission Operations Concept} \label{sec:experiment-pipeline-description}

Our ISRU mission includes a peg-in-hole teleoperated task, in which the MR collects a metal parallelepipedon and places it in a 3D-printed slot, provided with a hole of the same size as the sample (tolerance: \SI{0.002}{\meter}). During the whole experiment, the robot is placed outdoors and communicates remotely with the indoor ground station, hence the human operator can control the arm only via quasi-real-time visualization and teleoperations, monitoring the actual state of the mission through visual feedback consisting of the cameras' streams and 3D reconstruction of the robot state, both displayed in RVP.

The proposed mission operations concept consists of the following phases, illustrated in \figref{fig:experimental-pipeline}:
\begin{enumerate}
\item \textit{Pre-collection}: the MR moves in the proximity of the sample to collect. In our concept, it suffices that the sample is framed in the rear cameras, without a precise knowledge of its pose. Haptic teleoperations compensate for inaccuracies at the next step thanks to the visual feedback from the rear camera (an example is shown in \figref{fig:grasp-camera}). At the control station, this phase is supported by RVP, which assists the operator in planning collision-free trajectories. \label{itm:pre-grasp-pose}
\item \textit{Collection}: the MR precisely moves to the actual sample location. This step is performed via HCS: the human operator drives the robot towards a configuration where the gripper's fingers can safely close to collect the resource. 1:1 haptic position mapping allows for natural and accurate placement in a neighborhood of the resource, while the force feedback allows the user to feel the actual contact with the resource, thus inherently yielding a more reliable grasp and a safer motion. This phase ends when the MR closes the gripper's fingers, actually grasping the sample. \label{itm:grasp-pose}

\begin{figure}[!t]
    \centering
    \subfloat[Rear camera framing the sample in the collection phase \label{fig:grasp-camera}]{\includegraphics[width=0.475\columnwidth]{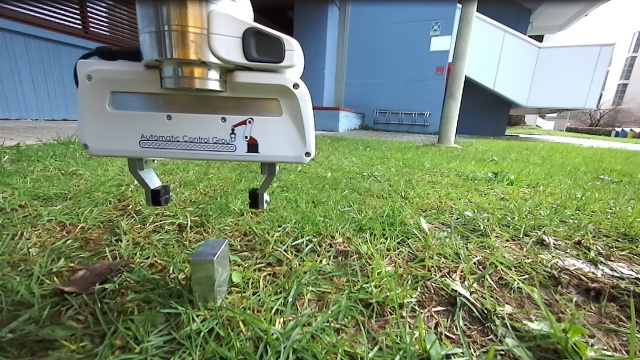}}
    \hspace{0.025\columnwidth}
    \subfloat[Front camera framing the slot in the utilization phase \label{fig:assembly-camera}]{\includegraphics[width=0.475\columnwidth]{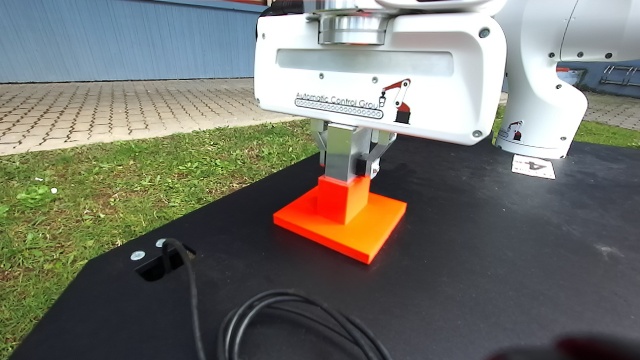}}
    \caption{Onboard camera images during teleoperations}
    \label{fig:camera-images}
\end{figure}

\item \textit{Post-collection}: the MR retracts from the soil (or fetching location). In real scenarios, this phase possibly requires the ER to navigate toward a different location for resource exploitation. \label{itm:grasp}
\item \textit{Pre-utilization}: the MR moves in the proximity of the location where resource utilization can happen, which possibly is in a different region of the MR's workspace. Thus, workspace extension is necessary, while human expertise is not required. The desired workspace extension is obtained by means of autonomous collision-aware planning through RVP.\label{itm:pre-assembly-pose}
\item \textit{Utilization}: the MR inserts the resource in its assembly slot. Here, human expertise is crucial for safe and accurate assembly. 1:1 position mapping, implemented by HCS, with arm visualization from both cameras and the 3D model, implemented by RVP, help the operator guide the robot near the hole (as in \figref{fig:assembly-camera}). Then, by sensing the force feedback on the haptic device, they can have a natural and comfortable feeling of the hole's walls, which allows for completing insertion with ease. \label{itm:peg-in-hole}
\item \textit{Post-utilization}: the MR opens the gripper's fingers, releasing the sample inside its assembly slot and achieving the mission goal. \label{itm:place}
\end{enumerate}

\subsection{Haptic Control System} \label{sec:haptic-control-system}

In teleoperations through haptic devices, various scaling and mapping techniques have been proposed in the literature \cite{radi_workspace_2012}, such as position control and rate control. In rate control, the displacement of the haptic device is interpreted as a velocity command, while position control consists of some linear position mapping from the haptic device's tip to the robot's end-effector.

In our design for ISRU, we connect the master and slave devices with 1:1 position mapping: this particular choice is especially important for the peg-in-hole task, as it yields the most direct, simple, and accurate transfer from human commands to robot motion. The linear mapping undergoes transformations to provide the user with the feeling of operating the MR from the ER cameras. Therefore, all displacements in the haptic device's workspace are relatively mapped in either rear or front camera frames (depending on the mission phase), so as to increase the naturalness of teleoperated control. In our design, the drawbacks of 1:1 position mapping, \ie a limited workspace and slow motions, are compensated by the execution of autonomous trajectories.

Concerning forces, we adopt linear mapping to transfer the force sensed at the robot's end-effector, in the order of tens of Newtons, to the haptic device, so as to be in the order of units. This is crucial to naturally and successfully perform a contact-rich manipulation task.

HCS also features gravity compensation of the robot's payload, as well as filtering of human tremors to generate stable poses for the robot and increase accuracy.

\subsection{Robot Visualization \& Planning} \label{sec:robot-visualization-for-remote-planning-and-motion-control}

\begin{figure}[!t]
    \centering
    \subfloat[Plan from initial configuration to fetching location \label{fig:pre-grasp-plan}]{\includegraphics[width=0.475\columnwidth]{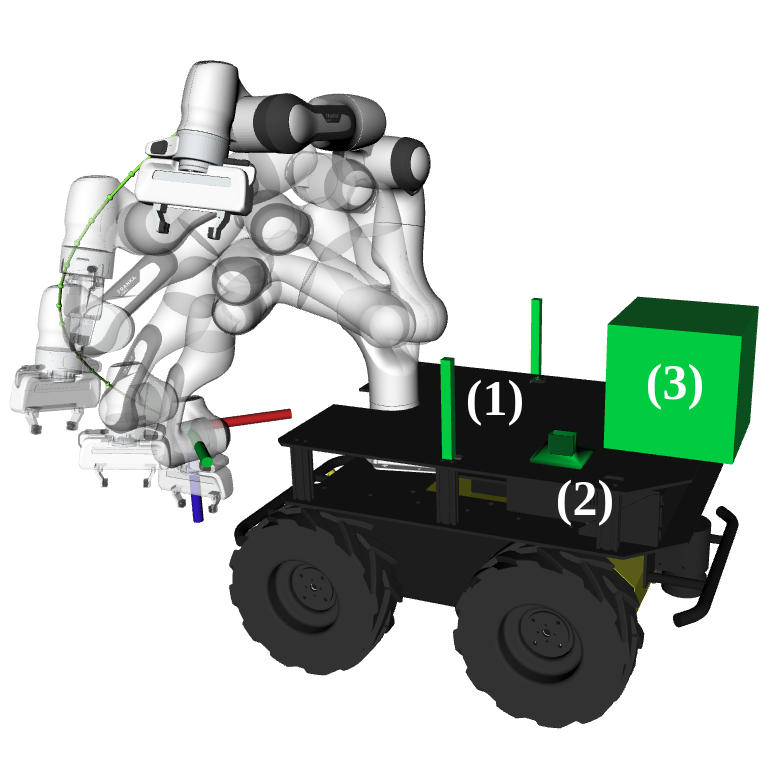}}
    \hspace{0.025\columnwidth}
    \subfloat[Plan from fetching location to assembly location \label{fig:pre-assembly-plan}]{\includegraphics[width=0.475\columnwidth]{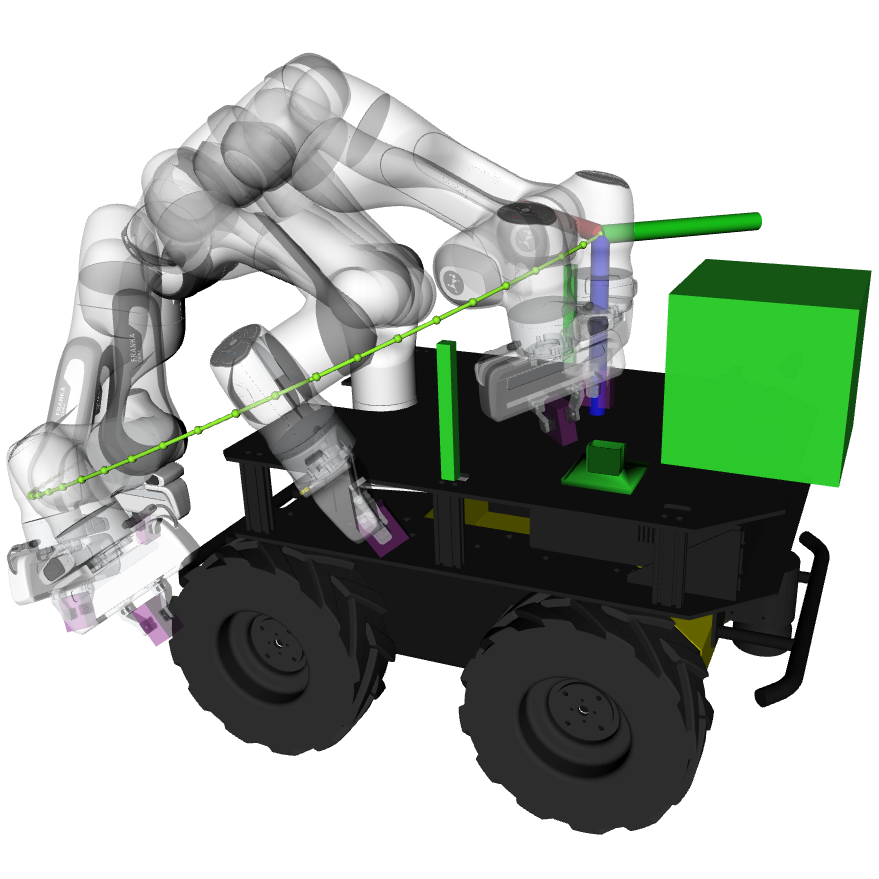}}
    \caption{Point-to-point trajectory planning in RVP. Together with ER and MR meshes, additional collision objects (in light green) are added to replicate (1) the ER's antennas, (2) the hole in which the sample is assembled, and (3) the front camera, modeled as a box to account for all possible orientations (being the front camera pair mounted on a passive pan-tilt unit, without encoders).}
    \label{fig:p2p-trajectory-planning}
\end{figure}

\begin{figure}[!t]
    \centering
    \subfloat[Engagement at the fetching location \label{fig:haptic-engagement-grasp}]{\includegraphics[width=0.475\columnwidth]{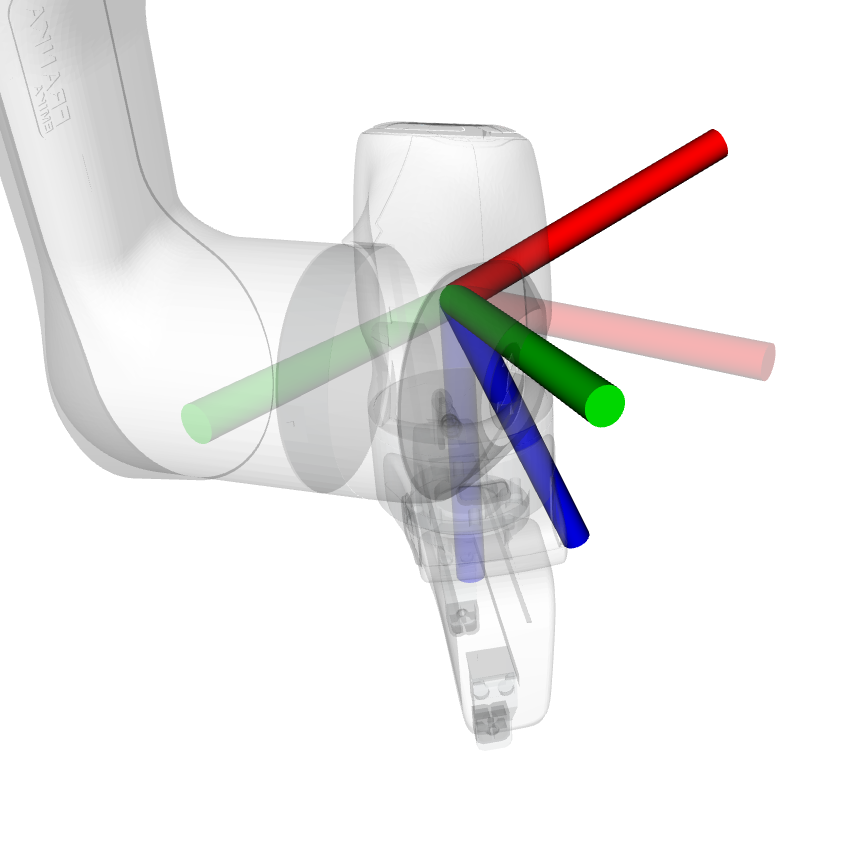}}
    \hspace{0.025\columnwidth}
    \subfloat[Engagement at the assembly location \label{fig:haptic-engagement-assembly}]{\includegraphics[width=0.475\columnwidth]{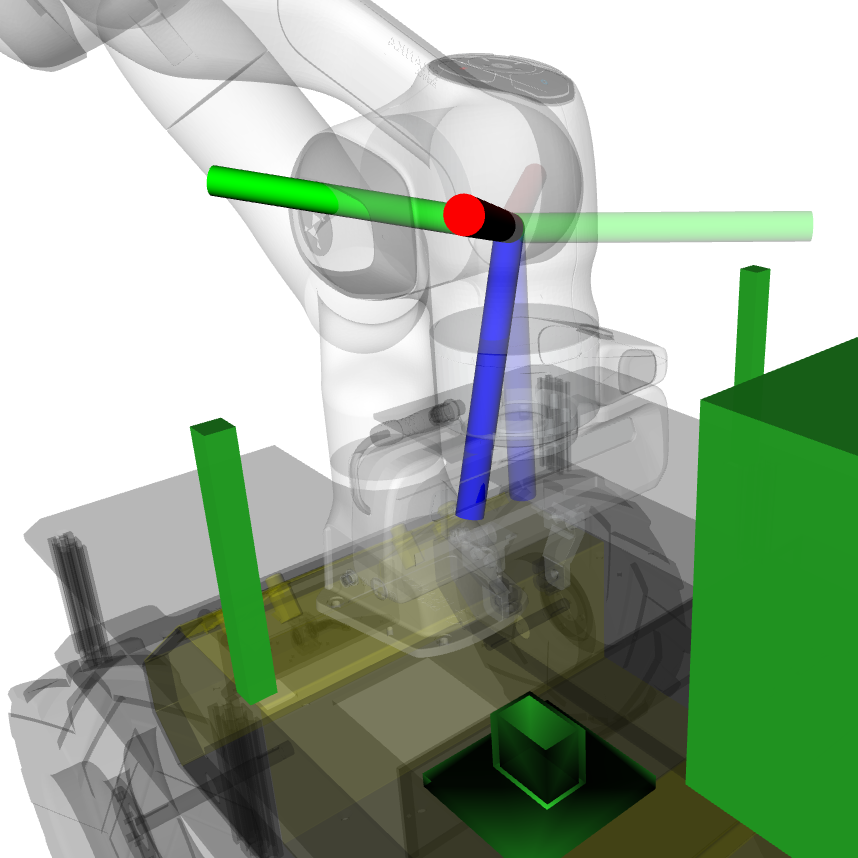}}
    \caption{Engagement procedure in RVP. The haptic device moving frame is shown with solid colors, while the fixed robot's end-effector frame is displayed as shaded.}
    \label{fig:haptic-engagement}
\end{figure}

RVP is designed to accomplish three goals:
\begin{enumerate}
\item Visualize and assess the robot state, \ie its joint configuration, in real-time. If the robot state is continuously updated and replayed in a 3D scene, the operator can assess the results of the commands, remotely sent from the control station. For complete awareness, cross-verification with camera streams is also possible through the RVP GUI. \label{itm:rviz-visualize-robot-state}
\item Plan, rehearse, and validate point-to-point collision-aware trajectories during pre-collection (see Section \ref{sec:experiment-pipeline-description}, step \ref{itm:pre-grasp-pose}) and pre-utilization phases (see Section \ref{sec:experiment-pipeline-description}, step \ref{itm:pre-assembly-pose}). Eventually, the planned trajectories (depicted in Figure~\ref{fig:p2p-trajectory-planning}) are uplinked to the robot. This is the feature that allows extending the teleoperated robot's workspace: indeed, differently from \cite{schmaus_realizing_2022}, our HRI, together with assisting the operator by providing visual feedback, allows combining accurate teleoperation positioning with a large workspace.\label{itm:rviz-plan-p2p-traj}
\item Manage the transitions between autonomous trajectories and teleoperations, \ie support the operator in engaging the MR with the haptic device when its stylus' and the robot's end-effector's orientations match, as visible in Figure~\ref{fig:haptic-engagement}. This component is crucial to activating teleoperations at different workspace locations with ease; indeed, without a visual assistive tool, the engagement would be practically unfeasible.\label{itm:rviz-engagement}
\end{enumerate}

\section{Results} \label{sec:results}

This section reports the results of rehearsing the operations of Section \ref{sec:experimental-setup} with the proposed HRI. The end-to-end system is shown deployed in a remote ISRU mission in \cite{ferrentino_autonomous_2022}, while this section's focus is on performing a formal assessment of different aspects of the system. With the aim to confirm the impact of haptic control on mission operations, in Section \ref{sec:assessment} we compare trials with force and visual feedback with trials with visual feedback only. In Section \ref{sec:analysis-of-the-system-in-case-of-communication-delays}, we assess the end-to-end system in case of communication delays. Such trials are performed by the same experienced operator on the same day. The operator has previous experience with haptic teleoperations for robotic interaction tasks and interaction control in general and a deep knowledge of how to operate the MR. 

Finally, in Section \ref{sec:ease-of-training-new-operators}, we assess the ease of learning to use the system to achieve the intended goal. These trials are therefore performed by operators with some knowledge of robotics, yet without any expertise in either the specific task or haptic teleoperations in general.

All the trials are conducted in a controlled environment, \ie a laboratory room (yet with the robot not in the view of the operator), with the workstation connected to the ER's wireless network. For each trial, the parallelepipedon to collect (see \figref{fig:grasp-camera}) is approximately placed in front of the camera, with little variations of the samples' pose across the trials.

\subsection{Haptic control assessment} \label{sec:assessment}

The task to perform is composed of two human-driven sub-tasks (\ie, \textit{Fetching} and \textit{Assembly}), both requiring haptic teleoperations. With reference to the pipeline presented in Section \ref{sec:experiment-pipeline-description}, \textit{Fetching} includes steps \ref{itm:grasp-pose}--\ref{itm:grasp}, while \textit{Assembly} corresponds to steps \ref{itm:peg-in-hole}--\ref{itm:place}. \textit{Fetching} is considered successful if the object is stably grasped, and the arm is lifted with the sample firmly held by the fingers. \textit{Assembly} is considered successful if the sample is assembled in its slot, and the fingers are open. If, in any sub-task, the measured forces exceed a safety threshold (above which the sample is considered as damaged), the MR controller stops and the \textit{Task} is counted as failed.

We exercise the whole operational procedure 20 times excluding the haptic feedback (Scenario A), and 20 times including the haptic feedback (Scenario B); in the former case, the user can solely rely on visual feedback. The overall results are reported in Table \ref{tab:results} in terms of the success rate of the \textit{Fetching} and \textit{Assembly} sub-tasks.

\begin{table}
\centering
\caption{Success rate of the mission operations out of 20 trials in different scenarios}
\label{tab:results}
\begin{tabular}{c|c|c|c|c}
Scenario & Force feedback & Delay $d$           & Fetching & Assembly \\ 
\hline
A        & $\xmark$       & $\SI{0}{\second}$   & 85\%       & 40\% \\
B        & $\cmark$       & $\SI{0}{\second}$   & 95\%       & 90\% \\
C        & $\cmark$       & $\SI{0.5}{\second}$ & 95\%       & 90\% \\
D        & $\cmark$       & $\SI{1.0}{\second}$ & 95\%       & 65\%
\end{tabular}
\end{table}

The results show that considerable improvements are achieved when perceiving the forces: indeed, the \textit{Assembly} success rate is increased by a factor of 2.25, while sensible improvements (the success rate is 1.06 times higher) are observed for \textit{Fetching} too, hence almost nullifying the probability of a failure, highlighting how the implicit compliance delivered by the human-in-the-loop is fundamental for the sample integrity.

\subsection{Analysis of the system in case of communication delays} \label{sec:analysis-of-the-system-in-case-of-communication-delays}
Since we aim at replicating a planetary manipulation task, we test the robustness of our system by introducing, via software, a delay in the communication link between the MR and the workstation. Given a delay $d$, if an operator forwards a command by moving the haptic device at time $t$, the MR receives the reference at time $t + d$, and the resulting forces produced by the motion are fed back to the operator at approximately $t + 2d$.

We perform 20 experiments with $d = \SI{0.5}{\second}$ (Scenario C), and 20 experiments with $d = \SI{1.0}{\second}$ (Scenario D), exercising, in both scenarios, the procedure detailed in Section \ref{sec:experiment-pipeline-description}, as before. We assume no inherent delay in the communication, besides the one introduced via software. The operator can rely on both visual and haptic feedback to perform the task, with the state of the robot displayed by RVP.

The results of the trials in Scenarios C and D are reported in Table~\ref{tab:results}. Although a round-trip delay of $2d = \SI{1.0}{\second}$, the performances of Scenario A are preserved. On the other hand, a remarkable decrease in performance is registered in Scenario D. In particular, because of the round-trip delay $2d = \SI{2.0}{\second}$, the success rate of the \textit{Assembly} phase drops from 90\% to 65\%, as the user's comfort during the peg-in-hole operation is degraded by the delayed perception of the contact forces between the sample and the assembly slot.

Although successful trials in Scenario D are twice more frequent than failures, a success rate of 65\% is not enough to consider the system's performance acceptable in the case of $d=\SI{1.0}{\second}$. Therefore, we consider $d=\SI{0.5}{\second}$ as the limit case, confirming the results of \cite{panzirsch_exploring_2022,schiele_towards_2015}.

\subsection{Ease of training new operators} \label{sec:ease-of-training-new-operators}

In order to assess the proposed HRI's ease of use in view of training new operators, we select three subjects with no experience with haptic teleoperations. Each participant is first explained procedures and tools, then they exercise the entire operational procedure several times. We consider the training to be complete when the operator is able to perform 5 consecutive successful trials.

The experiments foresee haptic feedback and no communication delays. We consider two metrics for the evaluation of ease of training, \ie the total training time and the number of attempts. The results are shown in Table \ref{tab:ease-of-training-results}.

\begin{table}
\centering
\caption{Training time of 3 different operators}
\label{tab:ease-of-training-results}
\begin{tabular}{c|c|c}
Operator & Number of attempts & Total training time \\ 
\hline
1        &      9       & \SI{1}{\hour} \SI{38}{\minute}\\
2        &      5       & \SI{35}{\minute}\\
3        &      10      & \SI{1}{\hour} \SI{20}{\minute}
\end{tabular}
\end{table}

Although the number of involved subjects is not enough to draw general conclusions, our preliminary results suggest that combining haptic teleoperations with suitable HRI and operational procedures might greatly simplify the training process for completely inexperienced operators. Therefore, we aim at extending our trials to a larger audience, possibly made of subjects with heterogeneous backgrounds.

\section{Conclusions} \label{sec:conclusions}

This work proposes a new concept of operations and associated HRI to assist human operators in the control of remote robotized systems for planetary ISRU missions. We adopt a haptic control system to allow for accurate remote manipulation, and a set of software tools and interfaces to plan and command trajectories from the control station. By alternating off-line planned trajectories and quasi-real-time teleoperations, the haptic control workspace is extended without sacrificing accuracy. At the same time, the proposed HRI preserves the system's ease of use.

Through the rehearsal across multiple trials of an ISRU scenario including a peg-in-hole manipulation task, we confirm that haptic control improves both safety and performance of the considered task, prevents damage to the collected sample, the robot, and its surroundings, increases human awareness and allows compensating for communication delays. In addition, our preliminary results suggest that inexperienced operators could be efficiently trained to complete the task with ease. This motivates further investigations involving a larger audience of operators with a heterogeneous background.

Through the employment of the proposed HRI, we assess the impact of the haptic feedback on mission success: an interaction task requiring an elevated degree of accuracy can be performed 2.25 times more successfully when compared to a system in which classical position-based teleoperations are adopted. Also, the system is robust to round-trip communication delays up to $\SI{1.0}{\second}$ and can be further improved by adopting more sophisticated state-of-art teleoperation techniques.

\section*{Acknowledgment}

The authors would like to thank Francesco Avallone for refining the controller implementation, making this work possible.

\bibliographystyle{IEEEtran}
\bibliography{metroaerospace-2023-haptic-teleop.bib}

\end{document}